\documentclass[pdflatex,sn-mathphys-num]{sn-jnl}

\usepackage{orcidlink}
\usepackage{csquotes}
\usepackage{ragged2e,booktabs,tabularx}
\usepackage{tabularray}
\usepackage{siunitx}
\usepackage{rotating}
\usepackage{multirow}
\usepackage{amsmath}
\usepackage{amssymb}
\usepackage{amsthm}
\usepackage{algorithm2e}
\SetKwComment{Comment}{/* }{ */}
\RestyleAlgo{ruled}
\usepackage{soul}
\usepackage{xcolor, color, soul}
\usepackage{lmodern}
\usepackage{subcaption}
\usepackage{enumitem}
\usepackage{mathtools, nccmath}
\usepackage[english]{babel}

\theoremstyle{definition}
\newtheorem{definition}{Definition}

\newcommand{\PreserveBackslash}[1]{\let\temp=\\#1\let\\=\temp}
\newcolumntype{C}[1]{>{\PreserveBackslash\centering}p{#1}}

\begin{document}

\title[Article Title]{Navigation Variable-based Multi-objective Particle Swarm Optimization for UAV Path Planning with Kinematic Constraints}

 \author[1,2]{\fnm{Thi Thuy Ngan} \sur{Duong} \orcidlink{0009-0005-9768-7084}} \email{nganduong@unist.ac.kr}

 \author[2]{\fnm{Duy-Nam} \sur{Bui}\orcidlink{0009-0001-0837-4360}} \email{duynam@ieee.org}
% % \equalcont{These authors contributed equally to this work.}

 \author*[3]{\fnm{Manh Duong} \sur{Phung}\orcidlink{0000-0001-5247-6180}} \email{duong.phung@fulbright.edu.vn}
% % \equalcont{These authors contributed equally to this work.}

 \affil[1]{\orgdiv{Department of Electrical Engineering}, \orgname{Ulsan National Institute of Science and Technology}, \orgaddress{\street{UNIST-gil 50}, \city{Ulsan}, \postcode{44919}, \country{Korea}}}

 \affil[2]{\orgdiv{Faculty of Electronics and Telecommunications}, \orgname{Vietnam National University}, \orgaddress{\street{144 Xuan Thuy}, \city{Hanoi}, \postcode{100000}, \country{Vietnam}}}

 \affil[3]{\orgdiv{Undergraduate Faculty}, \orgname{ Fulbright University Vietnam}, \orgaddress{\street{105 Ton Dat Tien}, \city{Ho Chi Minh City}, \postcode{700000}, \country{Vietnam}}}

% \affil[4]{\orgdiv{School of Electrical and Data Engineering}, \orgname{University of Technology Sydney}, \orgaddress{\street{15 Broadway}, \city{Ultimo NSW 2007}, \postcode{2148}, \country{Australia}}}


\abstract{
Path planning is essential for unmanned aerial vehicles (UAVs) as it determines the path that the UAV needs to follow to complete a task. This work addresses this problem by introducing a new algorithm called navigation variable-based multi-objective particle swarm optimization (NMOPSO). It first models path planning as an optimization problem via the definition of a set of objective functions that include optimality and safety requirements for UAV operation. The NMOPSO is then used to minimize those functions through Pareto optimal solutions. The algorithm features a new path representation based on navigation variables to include kinematic constraints and exploit the maneuverable characteristics of the UAV. It also includes an adaptive mutation mechanism to enhance the diversity of the swarm for better solutions. Comparisons with various algorithms have been carried out to benchmark the proposed approach. The results indicate that the NMOPSO performs better than not only other particle swarm optimization variants but also other state-of-the-art multi-objective and metaheuristic optimization algorithms. Experiments have also been conducted with real UAVs to confirm the validity of the approach for practical flights. The source code of the algorithm is available at \url{https://github.com/ngandng/NMOPSO}.
}


\maketitle

\section{Introduction}
Path planning is an essential problem for unmanned aerial vehicle (UAV) applications because it determines the flight path from the starting position to the destination that a UAV needs to follow to complete its mission \mbox{\cite{wang2023uav, puente2022review}}. The path should be optimal in certain criteria such as shortest length or minimal energy consumption. It also needs to meet constraints imposed by the kinematic model and safe operation of the UAV \cite{9014596}. Some objectives and constraints, however, may contradict leading to the non-existence of a single global optimal path. Path planning techniques therefore need to balance those requirements to obtain best possible solutions.

In the literature, A*, sampling-based algorithms, and artificial potential field (APF) methods are among the most popular for path planning \mbox{\cite{ZHAO201854}}. A* uses heuristics to guide its search in finding the shortest path between the start and goal positions \cite{mandloi2021unmanned, 9133589}. By maintaining a cost function originating from the start node, A* extends a path one edge at a time until the goal is reached. However, A* is limited in scalability as its discretization of the search space causes the number of cells to increase rapidly with the size of the space.

The APF method does not discretize the search space but defines it as a potential field formed by surrounding objects \cite{FAN2022105182,9538804,9234396}. The UAV is then modeled as a particle traveling in the field, attracted by the target and repelled by the obstacles. As the result, the UAV will move toward the goal along a smooth path while avoiding obstacles. The path generated, however, are not optimized and degraded when the complexity of the environment increases.

The sampling-based method, on the other hand, uses randomization to expand a tree representing the path until it reaches the goal position \cite{FAN2022105182, Sababha2022}. This approach guarantees to find a path to the goal if such a path exists. For example, the rapidly-exploring random trees (RRT) algorithm samples the search space randomly in a way biased toward the large unexplored areas \cite{kothari2013}. As time passes, the algorithm explores more areas and eventually finds the route to the goal. However, the RRT algorithm does not optimize the length of the path during the search process. Its variants such as RRT* \cite{5980479} can shorten the path but more computation is required.

Recently, nature-inspired optimization techniques have been used in UAV path planning due to their ability to produce optimal solutions \cite{9064786,Duan2014,reviewIETE}. These techniques use cost functions to formulate path planning as an optimization problem and then solve it with nature-inspired algorithms like the firefly algorithm (FA) \cite{CHENG2020132},  genetic algorithm (GA) \cite{Schacht2018,9222146}, artificial bee colony algorithm (ABC) \cite{XU2010535}, particle swarm optimization (PSO) \cite{PHUNG201725,9795684,SPSO,9145574}, and ant colony optimization (ACO) \cite{MORIN2022}. These algorithms consider a path as a candidate solution and then use swarm intelligence to improve it. The type of swarm intelligence varies depending on the phenomena the algorithm relies on. The GA uses mutation and crossover operators. The DE also uses mutation but combines it with differential evolution. On the other hand, the ACO uses pheromones and randomization to direct the search process. Other algorithms, such as the ABC, FA, and PSO utilize the social cognition behavior of a swarm to explore the solution space. 

Among nature-inspired algorithms, the PSO is often referred to as an efficient method capable of achieving optimal solutions with a high convergence rate. It is also less sensitive to initial conditions and can be adapted to various environment structures \cite{1216163,PSOEberhart}. The PSO obtains those features by balancing the personal experience of each individual and the experience of the whole swarm to find potential regions in the solution space. Several variants of PSO have been introduced for path planning like the discrete PSO (DPSO) \cite{PHUNG201725}, hybrid PSO \cite{9795684}, quantum-behaved PSO (QPSO) \cite{5941032}, and spherical vector-based PSO (SPSO) \cite{SPSO}. Most of the algorithms, however, are single objectives in which they combine objectives in a single cost function via a weighted sum. While this approach is simple to implement, combining multiple objectives leads to a cost function whose optimality does not represent the optimality of individual objectives. In practice, most objectives do not simultaneously reach their optimal values. Some of them are even contradicting that require a different approach called multi-objective optimization.

In the multi-objective optimization direction, several techniques have been introduced for path planning, such as the multi-objective firefly algorithm (MOFA) \cite{Hidalgo-Paniagua2017}, multi-objective PSO (MOPSO) \cite{9014596,ZHANG2013172}, and nondominated sorting genetic algorithm II (NSGA-II) \cite{Ahmed2013}. These algorithms use a similar mechanism called the non-domination principle to direct solutions toward Pareto optimal regions. In particular, NSGA-II uses a crowded-comparison operator with two attributes: non-domination rank and crowding distance to spread solutions across potential regions \cite{996017}. MOFA uses the search method from the Firefly algorithm to find non-dominated solutions and then carries out a migration procedure to maintain the diversity of the Pareto front \cite{Yang2013}. In MOPSO, non-dominated solutions are used to guide the exploration of the solution space. Two factors including a position update mechanism and a mutation operator are used to promote diversity \cite{2006MultiObjectivePSO}. Multi-objective optimization has the advantage of finding best possible solutions that form a set of non-dominated solutions called the Pareto front. Current algorithms, however, have not incorporated constraints imposed by the UAV kinematics into the search process. Since those constraints are essential to generating flyable paths, it is important to include them in the algorithm.

In this study, a new algorithm named navigation variable-based multi-objective PSO (NMOPSO) is proposed to generate flyable and Pareto-optimal paths for UAVs. First, a set of objective functions and navigation variables are defined to incorporate requirements and constraints for optimal UAV operation. The multi-objective PSO is then used to find a set of non-dominated solutions that best fit the objective functions. A mutation mechanism is introduced to avoid premature convergence and improve the search performance. Four scenarios were created based on real digital elevation model (DEM) maps of different complexity to evaluate the performance of the proposed approach. Our contributions in this work include (i) the proposal of the NMOPSO that includes kinematic constraints and multiple objectives to generate Pareto optimal paths for the autonomous operation of UAVs; (ii) the introduction of an adaptive mutation operator to improve the swarm performance; (iii) the derivation of a formula inspired by the Denavit-Hartenburg representation to convert navigation variables to Cartesian coordinates for efficient path searching and evaluation; (iv) the implementation of experiments with real UAVs to confirm the validity of the NMOPSO for practical operation.

The rest of this paper is structured as follows. Section \ref{sec:Problem} introduces the kinematic model and definition of objective functions. Section \ref{sec:moo} presents the proposed algorithm, NMOPSO. Section \ref{sec:result} provides comparison and experiment results. A conclusion is drawn in section \ref{sec:con} to end the paper.

\section{Problem Formulation}
\label{sec:Problem}
This section presents the kinematic model of the UAV and objective functions defined for the path planning problem.

\subsection{Kinematic model and constraints}
\label{sec:Kinematics}
Consider the UAV as a point moving in the environment. According to \cite{9199535}, its kinematic equations are described as follows:
\begin{equation}
    \left\{\begin{aligned}
        \dot{x}&=V\cos\alpha \cos\beta\\
        \dot{y}&=V\cos\alpha \sin\beta\\
        \dot{z}&=V\sin\alpha
    \end{aligned}\right.,
    \label{eqn:kinematic}
\end{equation}
where $[x,y,z]^T$ represents the position of the UAV; $V$ is the linear velocity; $\alpha$ and $\beta$  are respectively the climbing and turning angle. Due to physical limits, the velocity and angles of the UAV are subject to the following constraints: 
\begin{equation}
\label{eq:constraint}
    \left\{ \begin{array}{c}
    V_\text{min} \leq V \leq V_\text{max}\\
    \left\vert\Delta\alpha\right\vert=\left\vert\theta\right\vert\leq\theta_\text{max} \\ 
    \left\vert\Delta\beta\right\vert=\left\vert\psi\right\vert\leq\psi_\text{max}
    \end{array}\right.,
\end{equation}
where $V_\text{min}$ and $V_\text{max}$ are respectively the minimum and maximum linear velocities and $\psi_\text{max}$ and $\theta_\text{max}$ are respectively the maximum variations of turning and climbing angles. It is important to incorporate those constraints into the path planning algorithm so that feasible paths are generated for the UAV to follow.

\subsection{Objective functions for path planning}
\label{sec:objFunction}
Requirements for the path are formulated via the definition of objective functions. The functions are inspired by our previous work \mbox{\cite{SPSO}} but have been modified and normalized to the range $[0,1]$ to suit the multi-objective approach.

\subsubsection{Path length}
\begin{figure}
    \centering
    \includegraphics[width=0.5\textwidth]{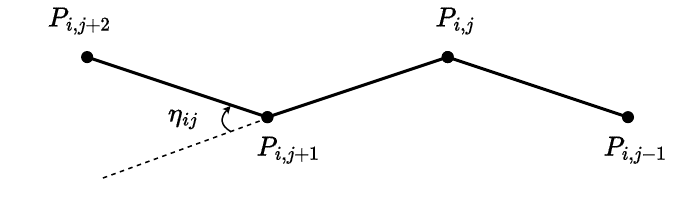}
    \caption{Illustration of a flight path and its variables}
    \label{fig:FlightPathVariables}
\end{figure}
In autonomous flights, a UAV is pre-loaded with a list of waypoints it needs to travel through. A flight path $p_i$ thus can be described by a set of $n$ waypoints, $p_{i}=\left\{P_{i1},...,P_{in}\right\}$, each includes three components, $P_{ij} = (x_{ij},y_{ij},z_{ij})$, as shown in Figure \ref{fig:FlightPathVariables}. Denote $\left\Vert\overrightarrow{P_{ij}P_{i,j+1}}\right\Vert$ as the Euclidean distance between two waypoints. The cost associated with the path length is then formulated as follows: 
\begin{equation}
    F_{1}= \begin{cases}
        1-\dfrac{\left\Vert\overrightarrow{P_{i1}P_{in}}\right\Vert}{\sum_{j=1}^{n-1} \left\Vert\overrightarrow{P_{ij}P_{i,j+1}}\right\Vert}, &\text{if }\left\Vert\overrightarrow{P_{ij}P_{i,j+1}}\right\Vert \geq R_\text{min}\\
        \infty, &\text{otherwise}\\
    \end{cases}
    \label{eqn:F1}
\end{equation}
where $R_\text{min}$ is the minimum distance between two waypoints. Finding the shortest path is equivalent to finding $p_i$ that minimizes $F_1$ in (\ref{eqn:F1}).

\subsubsection{Collision avoidance}

\begin{figure}
    \centering
    \includegraphics[width=0.35\textwidth]{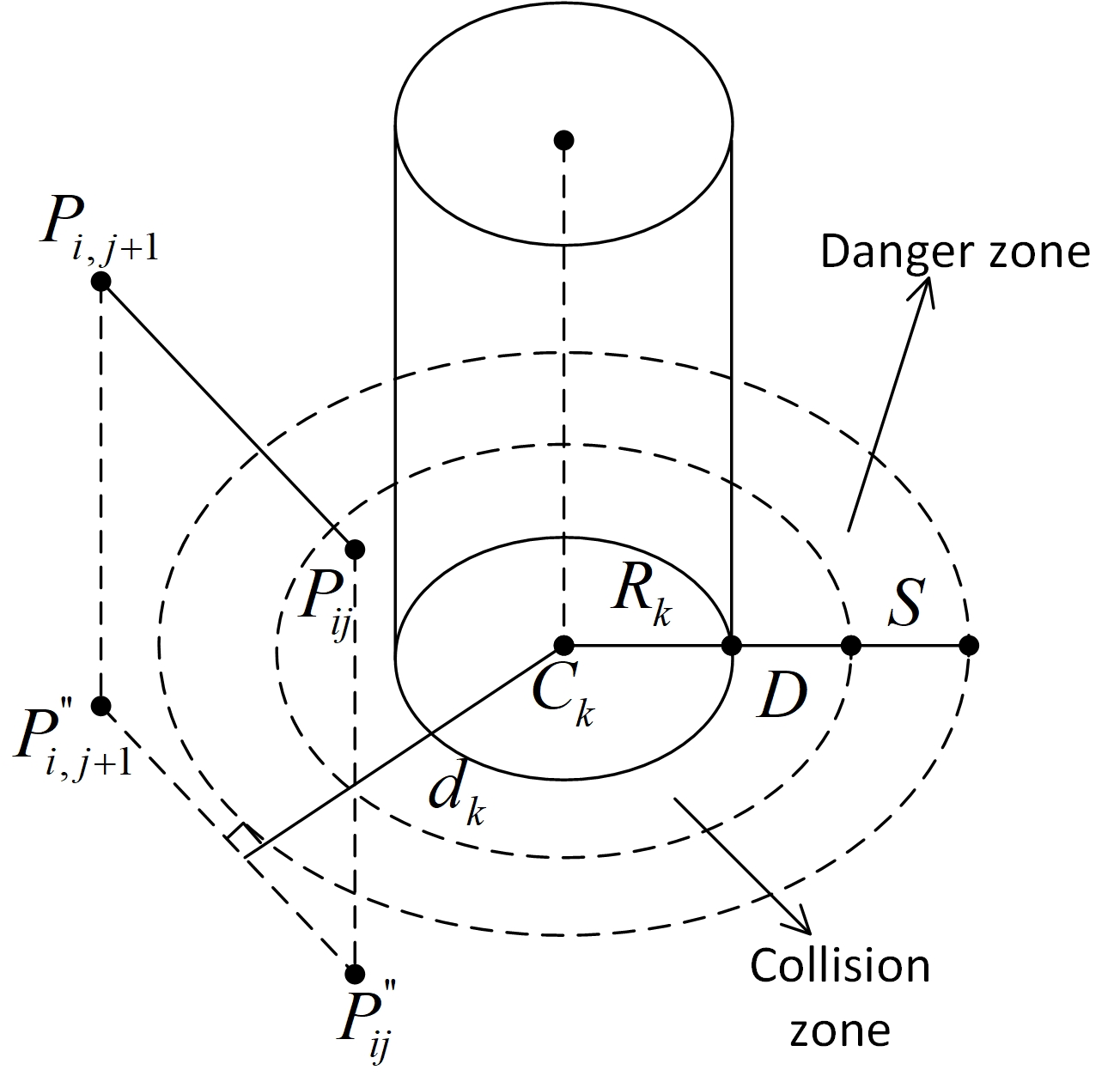}
    \caption{Obstacle avoidance}
    \label{fig:cost2}
\end{figure}
A flyable path should guide the UAV to avoid obstacles. In this work, an obstacle $k$ is modeled as a cylinder with center $C_k$ and radius $R_k$, as shown in Figure \ref{fig:cost2}. Let $d_k$ be the distance from the center of obstacle $k$ to path segment $P_{ij}P_{ij+1}$, $D$ be the UAV size, and $S$ be the safe distance from the UAV to the obstacle. The objective function for safe operation of the UAV is expressed as follows:
% \textcolor{red}{doi index $k$ sang vd $l$}

\begin{equation}
F_{2}=\frac{1}{K(n-1)}\sum_{j=1}^{n-1}\sum_{k=1}^{K}\mathcal{T}_{k}\left(\overrightarrow{P_{ij}P_{ij+1}}\right),
\label{eqn:F2}
\end{equation}
where $K$ is the number of obstacles in the working area and $\mathcal{T}_{k}$ is calculated as:
\begin{equation}
\label{eqn:ColisionT}
    \mathcal{T}_{k}\left(\overrightarrow{P_{ij}P_{ij+1}}\right)=\begin{cases}
    0, &\text{if }d_{k}\geq D+R_{k}+S \\
    1-\dfrac{d_k-D-R_k}{S}, &\text{if }D+R_{k}<d_{k}< D+R_k+S\\
    \infty, &\text{otherwise}
    \end{cases}
\end{equation}

Equation \eqref{eqn:ColisionT} implies that if a path segment is outside the danger zone (see Figure \ref{fig:cost2}), it incurs no additional cost to the objective function. However, if a path segment falls within the danger zone, its cost is inversely proportional to its distance to the obstacle. In case the path segment is within the collision zone, an infinite cost will be added to indicate a collision.

\subsubsection{Flight altitude}
During operation, it is preferable that the UAV flies at a stable altitude to minimize its energy consumption. Let $h_\text{max}$ and $h_\text{min}$ respectively be the maximum and minimum relative flight altitudes and $h_{ij}$ be the present altitude of the UAV. Let $h_\text{mean}=\dfrac{h_\text{max}+h_\text{min}}{2}$ be the preferable relative altitude for the flight. The objective function for the flight altitude is defined as: 
\begin{equation}
    F_{3}=\dfrac{1}{n}\sum_{j=1}^{n}\mathcal{H}_{ij},
    \label{eqn:F3}
\end{equation}
where
\begin{equation}
    \mathcal{H}_{ij}=\begin{cases}
    \dfrac{2\left\vert h_{ij}-h_\text{mean}\right\vert}{h_\text{max}-h_\text{min}}, & \text{if} \quad h_\text{min}\leq h_{ij}\leq h_\text{max}\\
    \infty, & \text{otherwise}
    \end{cases}
\end{equation}

\subsubsection{Smoothness}
Apart from maintaining the altitude, a flight path should also minimize variations in the turning angle of the UAV as they are directly proportional to energy consumption. As illustrated in Figure \ref{fig:FlightPathVariables}, the turning angle $\eta_{ij}$ is the angle between two consecutive path segments, $\overrightarrow{P_{ij}P_{i,j+1}}$ and $\overrightarrow{P_{i,j+1}P_{i,j+2}}$, and is computed as:

\begin{equation}
    \eta_{ij}=\arctan\left(\dfrac{\left\Vert\overrightarrow{{P_{ij}P_{i,j+1}}}\times{\overrightarrow{P_{i,j+1}P_{i,j+2}}}\right\Vert}{\overrightarrow{{P_{ij}P_{i,j+1}}}\cdot{\overrightarrow{P_{i,j+1}P_{i,j+2}}}}\right).
    \label{eta}
\end{equation}

Since this angle represents changes in the direction of the UAV along the flight path, the smooth cost $F_4$ is defined as:
\begin{equation}
    F_{4}=\dfrac{1}{n-2}\sum_{j=1}^{n-2}\dfrac{\left\vert\eta_{ij}\right\vert}{\pi},
    \label{eqn:F4}
\end{equation}
where $\pi$ is the maximum turning angle used for the normalization.

%%-------------Implement MOPSO----------------------------

\section{Multi-objective optimization approach} 
\label{sec:moo}
Given objective functions $F_1$ to $F_4$, an optimal path would simultaneously minimize all of them. Such a path, however, does not exist since those functions do not minimize at the same point. Some functions are even contradicting such that the decrease of one function leads to the increase of another. To overcome it, most studies combine those functions into a single objective function using a weighted sum \cite{SPSO,5941032}. While that approach is simple to implement, it is difficult to choose the right weight for each function and maintain the optimality of the obtained solution. In this work, we propose to use multi-objective optimization.

\subsection{Pareto-optimal solution}
Let $X$ be a path generated for the UAV. The path planning algorithm needs to find path $\hat{X}$ that simultaneously minimizes all cost functions $F_i$ defined in Section \ref{sec:objFunction},
\begin{equation}
  \hat{X} = \arg \min\left[F_1,F_2,F_3,F_4\right].
\end{equation}

According to the multi-objective optimization theory, there may not exist the optimal solution $\hat{X}$, but only the solution $X^*$ that is the best fit for all $F_i$. That solution is called the Pareto-optimal solution defined as follows:

\begin{definition}
  A solution $X \in \mathcal{X} \subset \mathbb{R}^4$ is \textbf{non-dominated} with respect to $\mathcal{X}$ if there is no other solution $X' \in \mathcal{X}$ such that $F_i(X') \le F_i(X)$ for every $i =1,..,4$ and $F_j(X') < F_j(X)$ for at least one $j$.
\end{definition}

\begin{definition}
  Let $\mathcal{F}$ be the feasible region. A solution $X^* \in \mathcal{F} \subset \mathbb{R}^4$  is considered \textbf{Pareto-optimal} if it is non-dominated with respect to $\mathcal{F}$.
\end{definition}

\begin{definition}
    The Pareto optimal set $\mathcal{P}^*$ is a set of non-dominated solutions defined by
    $\mathcal{P}^* = \left\{X^* \in \mathcal{F}| X^* \ \text{is Pareto-optimal} \right\}$.
\end{definition}

According to these definitions, a Pareto-optimal solution is one where no other solution can improve one objective without deteriorating at least one other objective. Unlike a single optimal solution, multiple Pareto-optimal solutions can exist, forming the Pareto optimal set or Pareto front. Several methods, such as weighting, lexicographic, and goal programming, can be used to find Pareto-optimal solutions \mbox{{\cite{Nyoman2018}}}. In this work, the particle swarm optimization method is chosen due to its efficiency and robustness \cite{4783028,trivedi2020simplified}.

\subsection{Muli-objective particle swarm optimization}
Particle swarm optimization (PSO) is a popular technique that was originally developed to solve single objective optimization problems using swarm intelligence. In PSO, a swarm of particles is first initialized so that the position of each particle represents a candidate solution. A cost function is then used to evaluate the fitness of those particles. Each particle of the swarm then moves in accordance to its best position and the best position of the swarm to improve its fitness until an optimal solution is obtained or the maximum number of iterations is reached. 

Let $x^t_i$ and $v^t_i$ be the position and velocity of particle $i$ at iteration $t$, respectively. Denote $x^t_{pbest,i}$ as the best position of particle $i$ and $x^t_{gbest}$ as the best position of the swarm at iteration $t$. The movement of particle $i$ is described by the following equations:
\begin{equation}
\begin{aligned}
    v^{t+1}_i &= wv^{t}_i + c_1r_1(x^t_{pbest,i} - x^t_i) + c_2r_2(x^t_{gbest} - x^t_i),\\
    x^{t+1}_i &= x^t_i + v^{t+1}_i
    \label{eqn:PSO}
\end{aligned}
\end{equation}
where $w$ is the inertial weight, $c_1$ and $c_2$ are respectively the cognitive and social coefficients, and $r_1$ and $r_2$ are random numbers drawn from the uniform distribution in the range of $[0,1]$.

When using PSO for multi-objective optimization, it is important to control the particles' distribution so that they can find non-dominated solutions. The particles should evolve under the guidance of non-dominated particles called leaders to spread across multiple potential regions. A popular approach is to define a repository to store non-dominated solutions and then use them as leaders \cite{1004388}. Each particle then selects a leader from the repository based on a crowd measure as follows.

\begin{figure}
    \centering
    \includegraphics[width=0.5\textwidth]{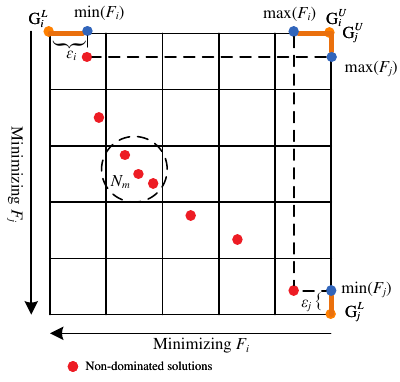}
    \caption{Illustration of a hypergrid and non-dominated solutions}
    \label{fig:hypergrid}
\end{figure}

Let $\mathcal{P}$ be the set of non-dominated solutions in the repository. A hypergrid is first established to allocate each particle in $\mathcal{P}$ to a hypercube as shown in Figure \ref{fig:hypergrid}. The coordinate of each particle is determined by their objective value \cite{rostami2016covariance}. Specifically, the lower bound, $G_i^L$, and upper bound, $G_i^U$, of the hypergrid in the dimension representing $F_i$ is determined as:
\begin{equation}
    G_i^L = \min_{x \in \mathcal{P}} F_i(x) - \epsilon_i
\end{equation}
\begin{equation}
    G_i^U = \max_{x \in \mathcal{P}} F_i(x) + \epsilon_i
\end{equation}
where $\epsilon_i$ is the padded grid length computed based on the number of grid divisions, $M$, as:

\begin{equation}
    \epsilon_i = \dfrac{1}{2(M-1)}\left(\max_{x \in \mathcal{P}} F_i(x) - \min_{x \in \mathcal{P}} F_i(x)\right).
\end{equation}
The coordinate of the hypercube containing particle $x$ in dimension $F_i$ is then computed by:
\begin{equation}
    c_i = \left\lfloor M\dfrac{F_i(x) - G_i^L}{G_i^U - G_i^L}\right\rceil,
    \label{eqn:GridLocation}
\end{equation}
where $\left\lfloor\cdot\right\rceil$ is the notation for rounding to the nearest integer. Let $N_m$ be the number of particles located in hypercube $m$. The crowd measure of that hypercube is computed as:
\begin{equation}
    \gamma_m = e^{-\kappa N_m},
\end{equation}
where $\kappa$ is a scaling coefficient. A leader for each particle of the swarm is then selected from the hypergrid in a random fashion with the selection probability proportional to the crowd measure:
\begin{equation}
    p_m = \dfrac{\gamma_m}{\sum_{l=1}^\mathcal{L}\gamma_l},
    \label{eqn:LeaderSelectionProb}
\end{equation}
where $\mathcal{L}$ is the number of hypercubes. Denote $x^t_{lbest,i}$ as the position of the selected leader for particle $i$. Equations for multi-objective PSO are written as: 
\begin{equation}
\begin{aligned}
    v^{t+1}_i &= wv^{t}_i + c_1r_1(x^t_{pbest,i} - x^t_i) + c_2r_2(x^t_{lbest,i} - x^t_i)\\
    x^{t+1}_i &= x^t_i + v^{t+1}_i.
\end{aligned}
    \label{MOPSO2}
\end{equation}
It can be seen that the main difference between the MOPSO and the original PSO is the replacement of $x^t_{gbest}$ by $x^t_{lbest,i}$ to diverge particle movements and look for multiple Pareto-optimal solutions. To further enhance the MOPSO, two improvements are introduced: one relates to the navigation variables, and the other involves a mutation mechanism.

\subsection{Navigation variables for particle position representation}
When using PSO for path planning, the position of each particle represents a candidate solution, which is a flight path. For path $p_i$ with $n$ waypoints $P_{ij} = [x_{ij},y_{ij},z_{ij}]^T$, position $X_i$ representing that path is expressed as:
\begin{equation}
\label{eq:pathX}
    X_i=(x_{i1},y_{i1},z_{i1},x_{i2},y_{i2},z_{i2},...,x_{in},y_{in},z_{in}).
\end{equation}
The use of Cartesian coordinates as in (\ref{eq:pathX}), however, does not incorporate maneuverable properties of the UAV into the path. It is therefore not efficient in finding non-dominated solutions or suitable to include kinematic constraints for feasible flight paths.  

Inspired by the operation of robot manipulators, we address this issue by considering a flight path as an articulated chain consisting of $n$ path segments. Each segment is described by a set of parameters similar to the Denavit–Hartenberg parameters \cite{4252158} including the climbing angle $\theta$, turning angle $\psi$, and length $r$. The end position of each segment then can be determined by the multiplication of transformation matrices representing the pose of previous path segments. 

\begin{figure}
    \centering
    \includegraphics[width=0.4\textwidth]{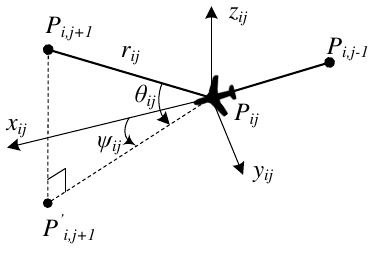}
    \caption{Illustration of variables for particle representation}
    \label{fig:angles}
\end{figure}

To implement this idea, at each waypoint $P_{ij}$, a coordinate frame attached to the UAV located at that point is defined, as shown in Figure \ref{fig:angles}. The $x$-axis points out of the UAV's front and is coincident with the line connecting $P_{i,j-1}$ and $P_{ij}$. The $y$-axis is directed out of the left side of the UAV and the $z$-axis is perpendicular to the $x$ and $y$ axes and is directed upward. Three new variables, $(r,\theta,\psi)$, are then defined as follows:

\begin{enumerate}[label=(\roman*)]

\item $r_{ij}$ is the distance between two consecutive nodes of path segment $\overrightarrow{P_{ij}P_{i,j+1}}$,  $r_{ij} = \left\Vert\overrightarrow{P_{ij}P_{i,j+1}}\right\Vert$;

\item $\psi_{ij}$ is the angle between two sequential path segments, $\overrightarrow{P_{i,j-1}P_{ij}}$ and $\overrightarrow{P_{ij}P'_{i,j+1}}$, where  $P'_{i,j+1}$ is the projection of $P_{i,j+1}$ on plane $Ox_{ij}y_{ij}$;

\item $\theta_{ij}$ is the angle between path segments $\overrightarrow{P_{ij}P_{i,j+1}}$ and  $\overrightarrow{P_{ij}P'_{i,j+1}}$.
% }
\end{enumerate} 

We call $(r,\theta,\psi)$ navigation variables. Position $\Gamma_i$ representing path $p_i$ in the navigation space then can be expressed as:
\begin{equation}
\label{eq:pathN}
    \left\{ \begin{array}{c}
        \Gamma_i=(r_{i1},\theta_{i1},\psi_{i1},r_{i2},\theta_{i2},\psi_{i2},...,r_{in},\theta_{in},\psi_{in})\\
    \left|\theta_{ij}\right|\leq\theta_{max} \quad \forall j \in \{1,..,n\} \\ 
    \left|\psi_{ij}\right|\leq\psi_{max} \quad \forall j \in \{1,..,n\}
    \end{array}\right.
\end{equation}
Different from Cartesian coordinates in (\ref{eq:pathX}), the use of navigation variables allows to include maneuverable properties of the UAV to the particles' position so that they can better explore the solution space. More importantly, kinematic constraints on the climbing and turning angles described in (\ref{eq:constraint}) can be directly included in the algorithm by limiting the range of those angles as in (\ref{eq:pathN}). As a result, the solution space is significantly narrowed down to increase the possibility of finding feasible and non-dominated solutions. 

During the optimization process, it is necessary to convert path $\Gamma_i$ in the navigation space (\ref{eq:pathN}) to its equivalent $X_i$ in the Cartesian space (\ref{eq:pathX}) to evaluate its fitness. This can be done by using transformation matrices. Specifically, the UAV's movement from waypoint $P_{ij}$ to waypoint $P_{i,j+1}$ can be divided into three sub-movements:
\begin{enumerate}[label=(\roman*)]
    \item Rotate an angle $\psi_{ij}$ about the $z_{ij}$-axis;
        %  \begin{equation*}
        %      R_z(\psi_{ij}) = \left[\begin{array}{cccc}
        %     \cos\psi_{ij} & -\sin\psi_{ij} & 0 & 0\\
        %     \sin\psi_{ij} & \cos\psi_{ij} & 0 & 0\\
        %     0 & 0 & 1 & 0\\
        %     0 & 0 & 0 & 1
        %     \end{array}\right]
        %  \end{equation*}
    \item Rotate an angle $\theta_{ij}$ about the $y_{ij}$-axis;
        % \begin{equation*}
        %      R_y(\theta_{ij}) = \left[\begin{array}{cccc}
        %     \cos\theta_{ij} & 0 & \sin\theta_{ij} & 0\\
        %     0 & 1 & 0 & 0\\
        %     -\sin\theta_{ij} & 0 & \cos\theta_{ij} & 0\\
        %     0 & 0 & 0 & 1
        %     \end{array}\right]
        %  \end{equation*}
    \item Translate $r_{ij}$ units along the $x_{ij}$-axis.
        % \begin{equation*}
        %     M_{x}(r_{ij}) = \left[\begin{array}{cccc}
        %     1 & 0 & 0 & r_{ij}\\
        %     0 & 1 & 0 & 0\\
        %     0 & 0 & 1 & 0\\
        %     0 & 0 & 0 & 1
        %     \end{array}\right]
        % \end{equation*}
\end{enumerate}
% \emph{Note: The positive of the rotation is counter-clockwise about the relative axis}
Hence, the transformation matrix between the frames located at two consecutive waypoints is computed as follows:
\begin{equation}
    T_{i,j+1}^{j}=R_z(\psi_{ij}) R_y(\theta_{ij}) M_{x}(r_{ij}),
 \label{eqn:TranspositionMatrix}
\end{equation}
where $R_y$ and $R_z$ are respectively the transformation matrices representing the rotation about the $y$ and $z$-axis, and $M_x$ is the translation matrix about the $x$-axis. 

Let $T_{iS}^{0}$ be the transformation matrix representing the start position and orientation of the UAV with respect to the inertial frame. The transformation matrix describing the position and orientation of frame $(Oxyz)_{ij}$ with respect to the inertial frame is given by:
\begin{equation}
    T_{ij}^{0}=T_{iS}^{0} T_{ij}^{S},
\end{equation}
where $T_{ij}^{S}$ is computed as:
\begin{equation}
    T_{ij}^{S}=T_{i1}^{S} T_{i2}^{1}\cdots T_{ij}^{j-1}.
\end{equation}
Finally, the Cartesian coordinates of waypoint $P_{ij}$ can be obtained from the navigation variables as:
\begin{equation}
    \tilde{P}_{ij}=T_{ij}^{0}\tilde{I},
    \label{eqn:CartPosition}
\end{equation}
where $\tilde{P}_{ij}=[P_{ij} \; 1]^T$ and $\tilde{I}=[0 \; 0 \; 0 \; 1]^T$.

\subsection{Region-based mutation}
When searching non-dominated solutions, certain particles of the swarm can be trapped in local optima, which affects the swarm's performance. For MOPSO, the chance of particles being trapped is higher due to their spread across multiple regions of the search space \cite{1004388}. We overcome this problem by introducing an adaptive mutation mechanism so that the level of mutation is proportional to the crowding distance of particles. For a random particle $x_{i}$ at iteration $t$, the mutation is conducted as follows:
\begin{equation}
    x_{ij}^t=x_{ij}^t+\mathcal{N}_{ij} G^t x^t_{pbest,i},
    \label{eqn:Mutation}
\end{equation}
where $j$ is a randomly selected component of $x_i$, $\mathcal{N}_{ij}$ is a random variable having the Gaussian distribution with zero mean and unit variance, and $G$ is a mutation gain.

To better fit the multi-objective optimization problem, $G$ is adjusted based on the distribution of non-dominated solutions in the Pareto optimal set. Since the spread of those solutions is proportional to the number of occupied hypercubes in the hypergrid, it can be used as a parameter to adjust $G$. When the number of occupied hypercubes is small, they indicate the dense concentration of particles over those regions. A large gain value is therefore needed to drive the particles to new regions. In contrast, when the particles are widely distributed, a small gain value will help to better explore those regions. Denote $N^t_r$ as the number of occupied hypercubes at iteration $t$. The mutation gain is determined as follows:
\begin{equation}
    G^t=\tanh\left(\dfrac{\Delta}{N_r^t}\right),
\end{equation}
where $\tanh\left(\cdot\right)$ represents the hyperbolic tangent and $\Delta$ is a pre-defined constant. This constant is set equal to the number of hypercubes occupied when the swarm is initialized.

\subsection{Implementation of the NMOPSO for UAV path planning}
\begin{algorithm}[!]
\caption{Pseudocode for the NMOPSO algorithm}\label{alg:MOSPSO}
    \Comment{Initialization}
    Get the search map and initial information \;
    Initialize swarm parameters $w$, $c_1$, $c_2$ \;
    Initialize grid dimension $M$, scaling coefficient $\kappa$, mutation coefficient $\Delta$ \;
    \ForEach{particle i in swarm}{
        Generate random path $\Gamma_{i}^{0}$ \;
        Get particle's position $X_{i}^{0}$ from $\Gamma_{i}^{0}$ \;
        Evaluate fitness $F_j(X_{i}^{0})$ of particle $i$ \;
        Set $pbest_{i}$ for particle $i$ based on the fitness \;
    }
    Initialize repository $\mathcal{P}$ \;
    Create a hypergrid and find grid location $c_i$ for solutions in $\mathcal{P}$ \tcc*[r]{Eq. \ref{eqn:GridLocation}}
    
    \Comment{Search}
    \While{not max\_iteration}
    {
        \ForEach{particle i in swarm}{
            Select a leader from $\mathcal{P}$ \tcc*[r]{Eq. \ref{eqn:LeaderSelectionProb}}
            Calculate velocity $v^t_i$ and update new position $\Gamma_{i}^{t}$ \tcc*[r]{Eq. \ref{MOPSO2}}
            Map $\Gamma_{i}^{t}$ to $X_{i}^{t}$ \tcc*[r]{Eqs. \ref{eqn:TranspositionMatrix}-\ref{eqn:CartPosition}}
            Update fitness $F_j(X_{i}^{t})$ \tcc*[r]{Eqs. \ref{eqn:F1}-\ref{eqn:F4}}
            Update $pbest_{i}$ \;
            Apply mutation \tcc*[r]{Eq. \ref{eqn:Mutation}}
        }
        Update $\mathcal{P}$\;      
        Update the hypergrid \tcc*[r]{Eq. \ref{eqn:GridLocation}}
    }
    Set Pareto optimal set $\mathcal{P}^* = \mathcal{P}$ \;
    Generate paths from non-dominated solutions in $\mathcal{P}^*$ \;
\end{algorithm}

The implementation of the NMOPSO is presented in Algorithm~\ref{alg:MOSPSO} together with the equations used. Compared to the original PSO, additional steps related to the hypergrid, leader selection, navigation variables, and mutation are added to find Pareto-optimal solutions. The algorithm stops when the maximum number of iterations is reached.

\section{Results} \label{sec:result}
To evaluate the performance of the proposed algorithm, a number of comparisons and experiments have been conducted with details below.

\subsection{Scenario setup}

Evaluations are conducted using real digital elevation model (DEM) maps of two areas on Christmas Island, Australia, each with distinct terrain structures \cite{https://doi.org/10.26186/89644}, as shown in Figures \ref{fig:PSOresultMap1} and \ref{fig:PSOresultMap2}. Each map is then augmented with obstacles to form four scenarios of different levels of complexity as shown in Figure \ref{fig:PSOresultMap2}. Comparisons are then carried out on these scenarios with pre-defined start and goal positions. Parameters of NMOPSO in all comparisons are chosen as follows: the number of path nodes $n = 10$; $w = 1$ with the damping rate of 0.98; $c_1 = 1.5$ and $c_2 = 1.5$; the number of grid divisions $M = 7$; the scaling coefficient $\kappa = 2$; the mutation coefficient $\Delta = 5$; and the turning and climbing angle limits $\theta_{max} = \psi_{max} = \pi/4$.

\subsection{Path planning results}

\begin{figure}
\begin{subfigure}[b]{.49\textwidth}
    \centering
    \includegraphics[width=\textwidth]{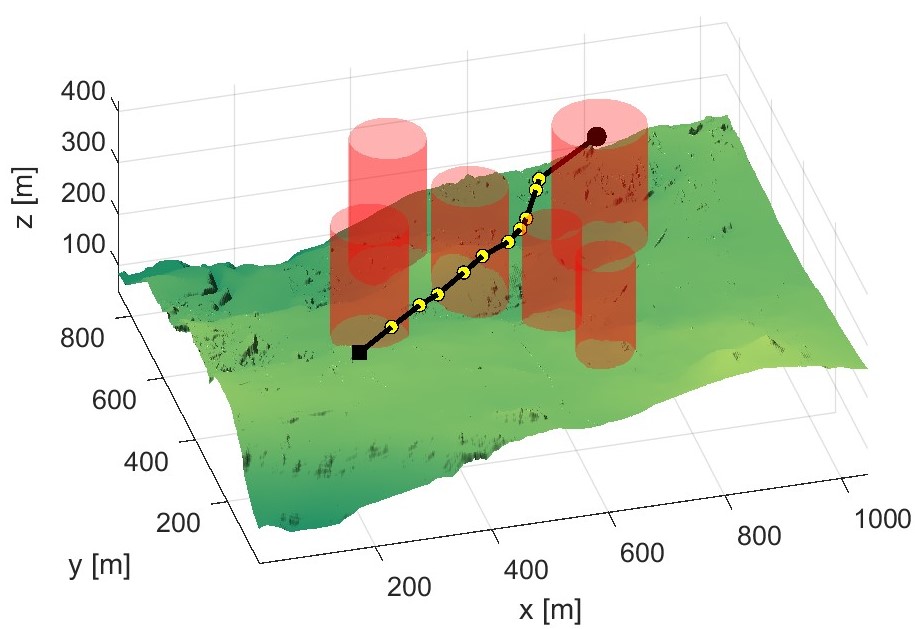}
    \caption{Scenario 1}
    \label{fig:model3xyz}
\end{subfigure}
\begin{subfigure}[b]{.49\textwidth}
    \centering
    \includegraphics[width=\textwidth]{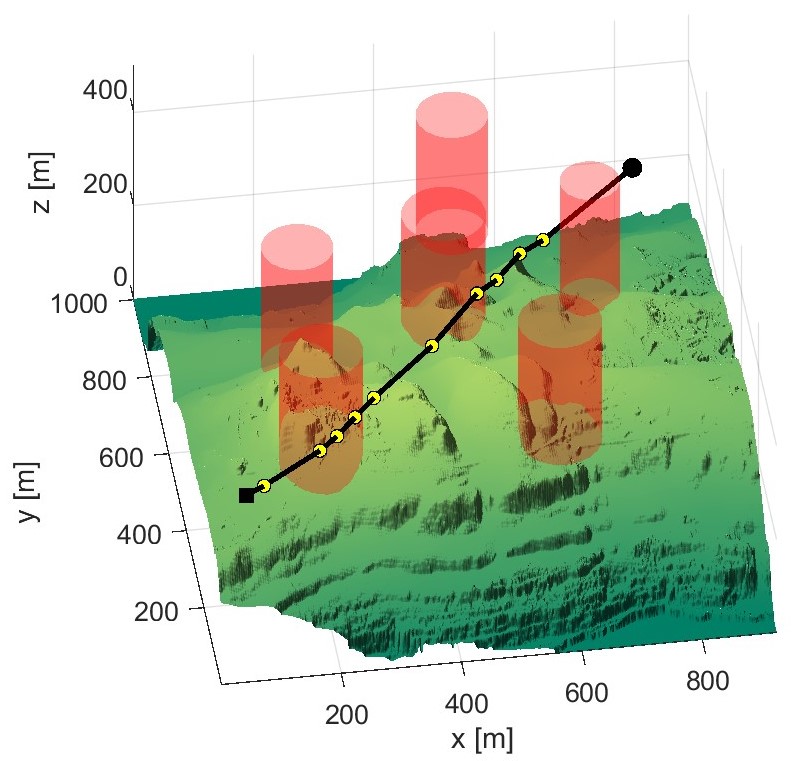}
    \caption{Scenario 4}
    \label{fig:model8xyz}
\end{subfigure}

\caption{3D view of the paths generated by the NMOPSO in scenarios 1 and 4}
\label{fig:model3NMOPSO}
\end{figure}

\begin{figure}
\begin{subfigure}[b]{.49\textwidth}
    \centering
    \includegraphics[width=\textwidth,height=3.5cm]{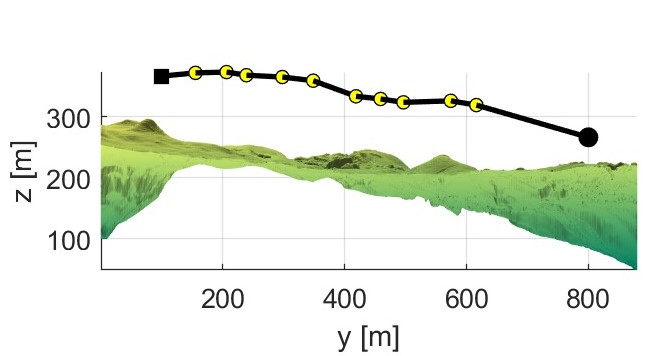}
    \caption{Scenario 1}
    \label{fig:model3z}
\end{subfigure}
\begin{subfigure}[b]{.49\textwidth}
    \centering
    \includegraphics[width=\textwidth,height=3.5cm]{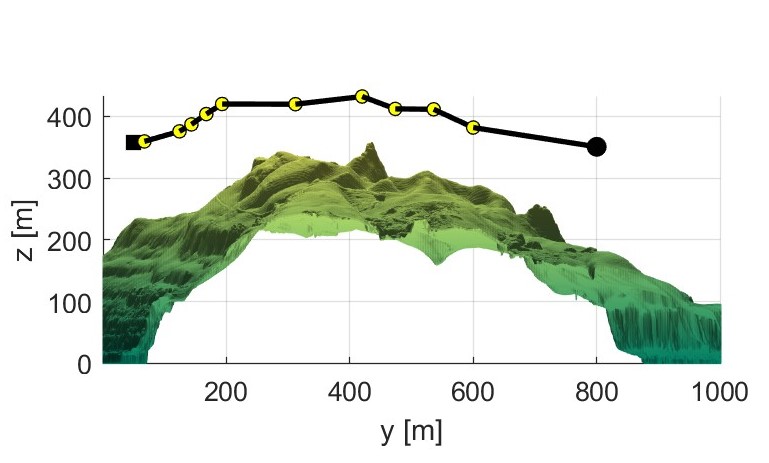}
    \caption{Scenario 4}
    \label{fig:model8z}
\end{subfigure}
\caption{Side view of the paths generated by the NMOPSO in scenarios 1 and 4}
\label{fig:model8NMOPSO}
\end{figure}

Figures \ref{fig:model3NMOPSO} and \ref{fig:model8NMOPSO} show the paths generated for scenarios 1 and 4. It can be seen that all paths are collision-free and successfully reach the goal positions. The side view of those paths shows that they adapt to the terrain structure but sharp changes in altitude are suppressed due to kinematic and smoothness constraints. The paths thus are feasible for the UAV to follow. Note that the paths shown in Figures \ref{fig:model3NMOPSO} and \ref{fig:model8NMOPSO} are just some among many non-dominated solutions in the Pareto Front obtained by the algorithm. Each solution dominates in certain objectives. Therefore, some non-dominated solutions may be preferred over others depending on the application.

\subsection{Comparison with other PSO variants}

\begin{figure}
\centering
    \begin{subfigure}[b]{.49\textwidth}
    \centering
    \includegraphics[width=\textwidth]{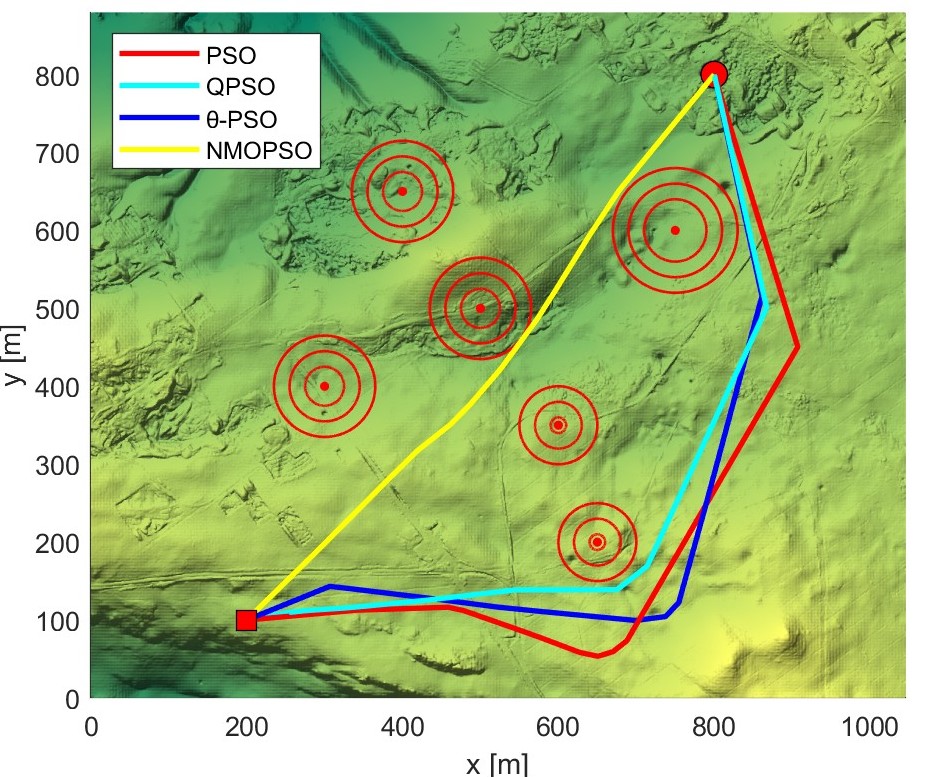}
    \caption{Scenario 1}
    \end{subfigure}
    \begin{subfigure}[b]{.49\textwidth}
    \centering
    \includegraphics[width=\textwidth]{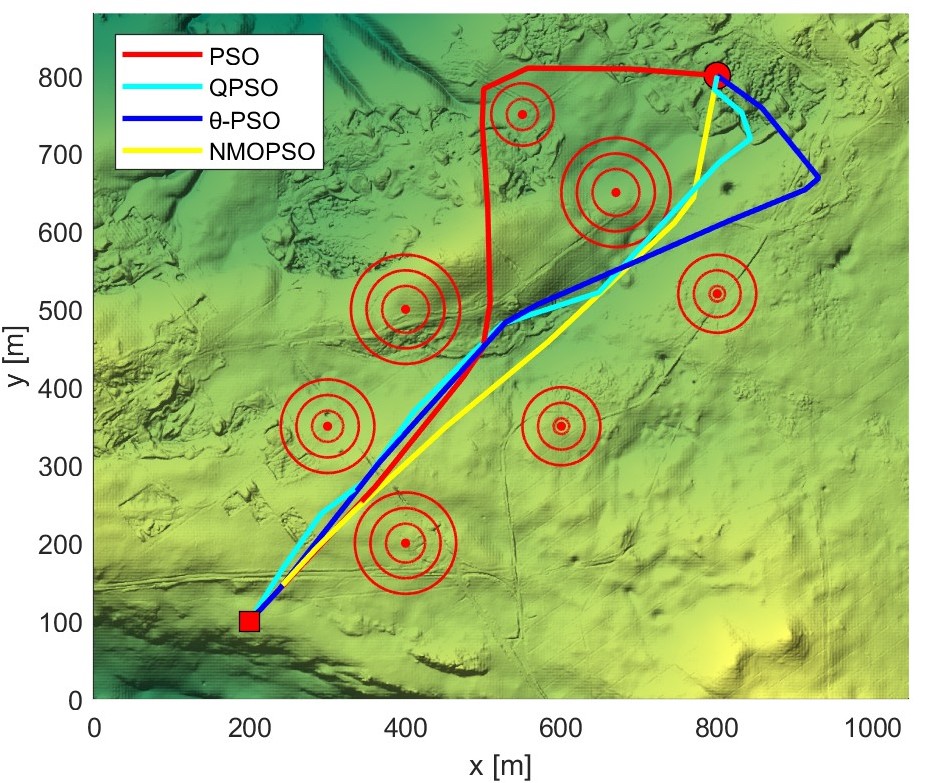}
    \caption{Scenario 2}
    \end{subfigure}
\caption{Top view of the paths generated by the PSO-based algorithms in simple elevation maps}
\label{fig:PSOresultMap1}
\end{figure}

\begin{figure}
\centering
    \begin{subfigure}[b]{.49\textwidth}
    \centering
    \includegraphics[width=\textwidth]{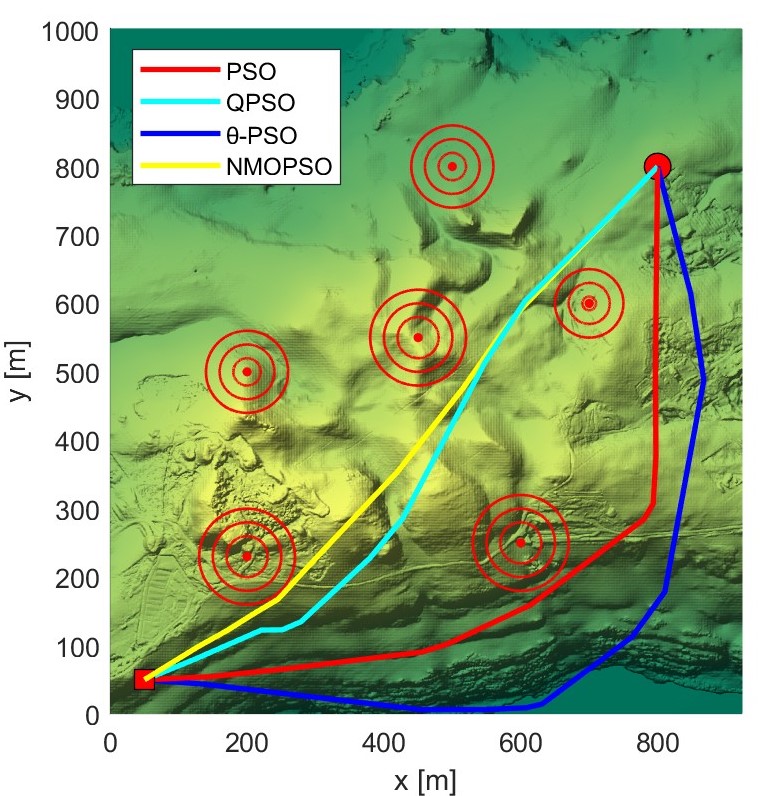}
    \caption{Scenario 3}
    \end{subfigure}
    \begin{subfigure}[b]{.49\textwidth}
    \centering
    \includegraphics[width=\textwidth]{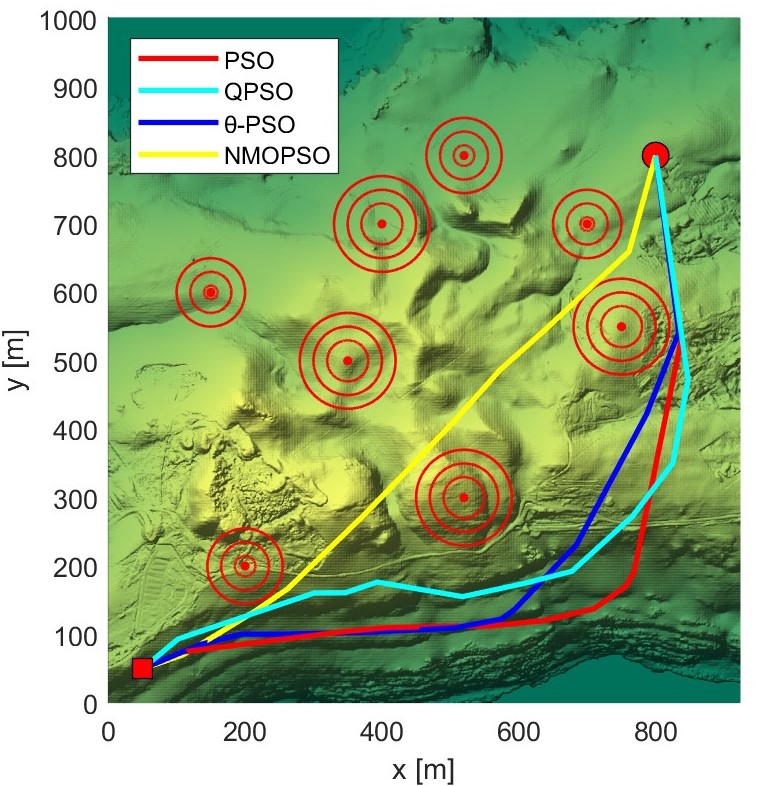}
    \caption{Scenario 4}
    \end{subfigure}
\caption{Top view of the paths generated by the PSO-based algorithms in complex elevation maps}
\label{fig:PSOresultMap2}
\end{figure}

% new table
\begin{table}
\caption{Comparison result with single-objective algorithms} 
\label{tbl:comparision_table1}
\centering
\begin{tabular}{C{1cm}C{1.2cm}C{1cm}C{1cm}C{1cm}C{1cm}C{1cm}C{1cm}C{1cm}C{1cm}C{1.2cm}}
\hline \hline
Scens. & Objective & \multicolumn{9}{c}{Algorithm} \\
          &           & \multicolumn{3}{c}{NMOPSO} & ABC      & DE         & GA         & PSO             & QPSO   & $\theta$-PSO          \\ \hline \hline
\multirow{4}{*}{1}         & F1        & \textbf{0.0127} & 0.0164              & 0.0382          & 0.1247   & 0.3991     & 0.2691     & 0.2709          & 0.2542 & 0.2782            \\
          & F2        & 0.0262          & 0.0254              & \textbf{0}      & 0.0005   & \textbf{0} & 0.0011     & \textbf{0}      & 0.0002 & 4.09E-9          \\
          & F3        & 0.289           & 0.086               & 0.0691          & 0.176    & 0.0154     & 0.2774     & \textbf{0.0003} & 0.1553 & 0.0004            \\
          & F4        & \textbf{0.036}  & \textbf{0.036}      & 0.0593          & 0.1272   & 0.0842     & 0.1932     & 0.0775          & 0.1065 & 0.0852            \\ \hline
\multirow{4}{*}{2}         & F1        & 0.0577          & 0.0348              & \textbf{0.0256} & 0.1767   & 0.3675     & 0.1311     & 0.1188          & 0.1224 & 0.1546            \\
          & F2        & \textbf{0}      & \textbf{\textbf{0}} & 0.0235          & 0.0003   & \textbf{0} & 0.0041     & 4.31E-5        & 0.0011 & 0.0002            \\
          & F3        & 0.1807          & 0.1855              & 0.3469          & 0.166    & 0.0207     & 0.3124     & \textbf{0.0032} & 0.1525 & 0.0218            \\
          & F4        & 0.0473          & 0.0487              & \textbf{0.0328} & 0.137    & 0.0891     & 0.1363     & 0.1152          & 0.1637 & 0.1199            \\ \hline
\multirow{4}{*}{3}         & F1        & 0.0261          & 0.0457              & \textbf{0.0189} & 0.2736   & 0.3081     & 0.2103     & 0.3167          & 0.1647 & 0.3502            \\
          & F2        & 0.0147          & \textbf{0}          & 0.0194          & 1.33E-6 & \textbf{0} & 0.0028     & 7.42E-7        & 0.0006 & \textbf{0}        \\
          & F3        & 0.0941          & 0.1185              & 0.1597          & 0.1666   & 0.0164     & 0.2662     & 6.55E-5        & 0.1186 & \textbf{5.86E-5} \\
          & F4        & 0.0505          & 0.0628              & \textbf{0.0356} & 0.1454   & 0.0871     & 0.2462     & 0.0858          & 0.1464 & 0.0867            \\ \hline
\multirow{4}{*}{4}         & F1        & 0.2133          & \textbf{0.0519}     & 0.1437          & 0.2706   & 0.234      & 0.3284     & 0.2962          & 0.2322 & 0.3034            \\
          & F2        & 0.0145          & 0.0059              & \textbf{0}      & 8.33E-6 & 1.28E-7   & \textbf{0} & 6.06E-5        & 0.0006 & 2.51E-7          \\
          & F3        & 0.0876          & 0.1605              & 0.0865          & 0.097    & 0.025      & 0.2747     & \textbf{0.0002} & 0.499  & 0.0005            \\
          & F4        & 0.0942          & \textbf{0.0745}     & 0.148           & 0.1151   & 0.0922     & 0.2178     & 0.0885          & 0.1298 & 0.0923          \\ \hline\hline  
\end{tabular}
\end{table}
%=========================================================
In this evaluation, the NMOPSO is compared with other single-objective PSO variants including the original PSO, quantum-behaved PSO (QPSO) \cite{QPSO}, and angle-encoded PSO ($\theta$-PSO) \cite{ThetaPSOuav}. Their parameters are chosen to be the same as the NMOPSO. However, the cost function is a weighted sum of the objectives as follows:

\begin{equation}
    F = \sum_{i=1}^{4} w_iF_i,
    \label{eqn:F}
\end{equation}
where $w_i$ is a weighting coefficient. In our implementation, $w_i=1$ since the objective functions are already normalized to $[0,1]$. 

Figures \ref{fig:PSOresultMap1} and \ref{fig:PSOresultMap2} show the paths generated by the algorithms. It can be seen that all algorithms are able to generate collision-free paths to the goal. The NMOPSO, however, produces the shortest and smoothest paths in all scenarios. This result can be further confirmed in Table \ref{tbl:comparision_table1}, which shows the values of the objective functions corresponding to the generated paths. It can be seen that the NMOPSO generates multiple Pareto optimal solutions. Some of them dominate certain objectives, while others have reasonable values across all objectives. They together outperform other PSO variants in most objectives.
\subsection{Comparison with other metaheuristic algorithms}

\begin{figure}
\centering
    \begin{subfigure}[b]{.49\textwidth}
    \centering
    \includegraphics[width=\textwidth]{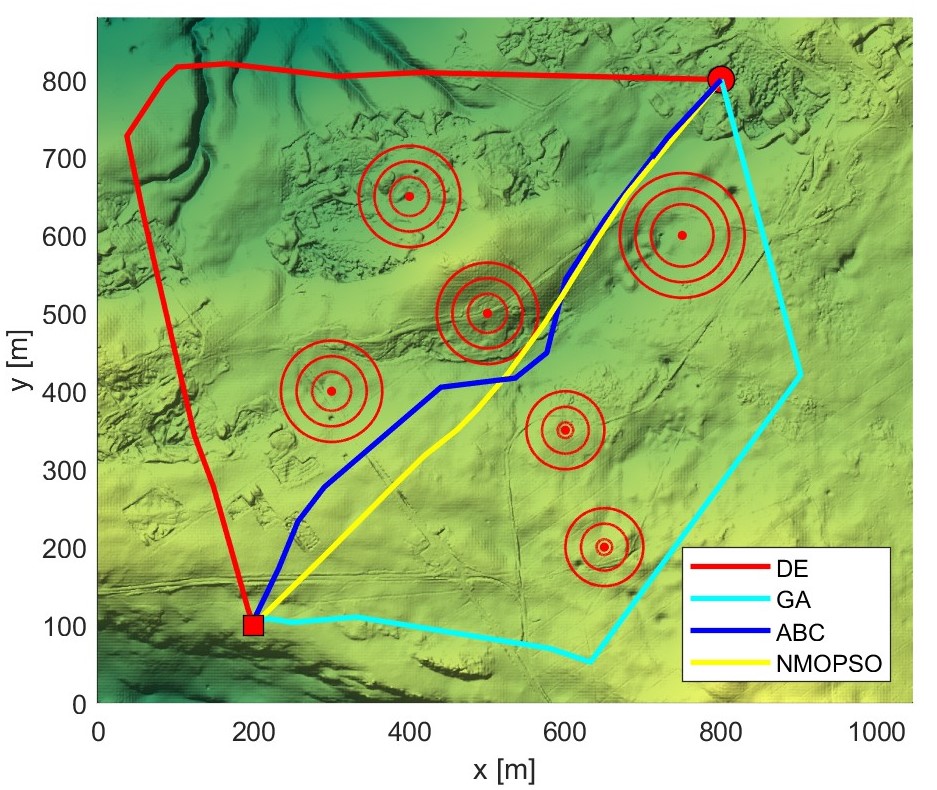}
    \caption{Scenario 1}
    \label{fig:meta1}
    \end{subfigure}%
    \begin{subfigure}[b]{.49\textwidth}
    \centering
    \includegraphics[width=\textwidth]{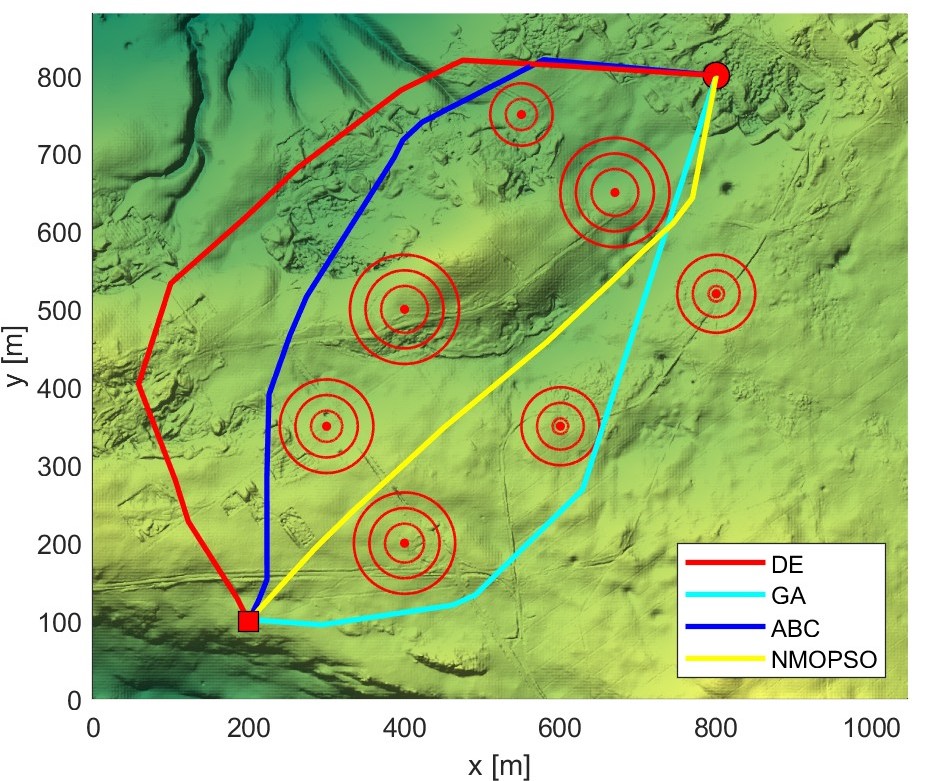}
    \caption{Scenario 2}
    \label{fig:meta2}
    \end{subfigure}

\caption{Top view of the paths generated by the NMOPSO and other meta-heuristic algorithms}
\label{fig:MetaheuristicALMap1}
\end{figure}

\begin{figure}
\centering
    \begin{subfigure}[b]{.49\textwidth}
    \centering
    \includegraphics[width=\textwidth]{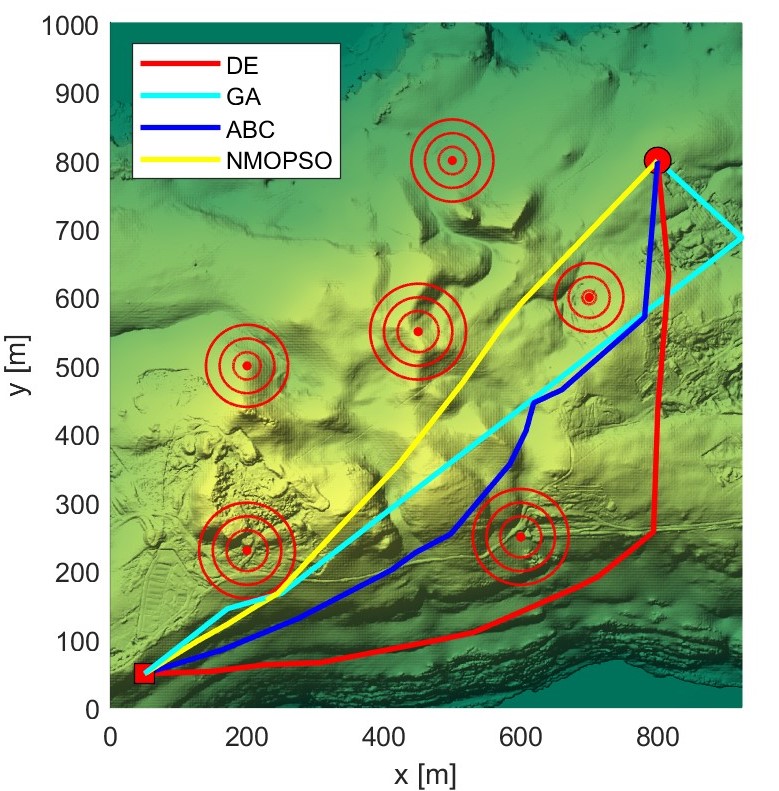}
    \caption{Scenario 3}
    \end{subfigure}
    \begin{subfigure}[b]{.49\textwidth}
    \centering
    \includegraphics[width=\textwidth]{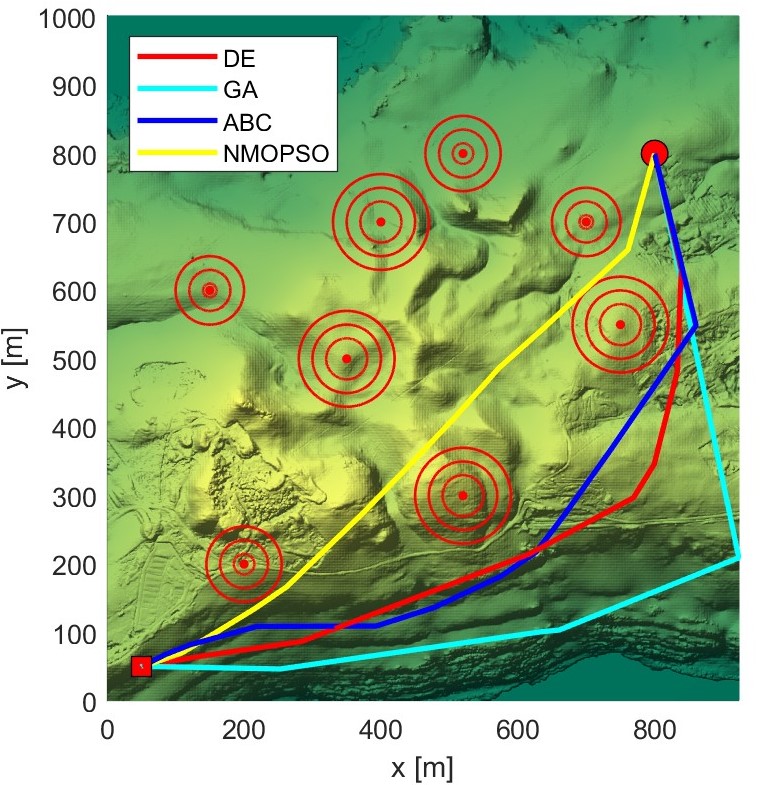}
    \caption{Scenario 4}
    \label{fig:otherAlgorithm4}
    \end{subfigure}
\caption{Top view of the paths generated by NMOPSO and other nature-inspired algorithms in complex elevation maps}
\label{fig:MetaheuristicALMap2}
\end{figure}

\begin{figure}
\begin{subfigure}[b]{.49\textwidth}
    \centering
    \includegraphics[width=\textwidth]{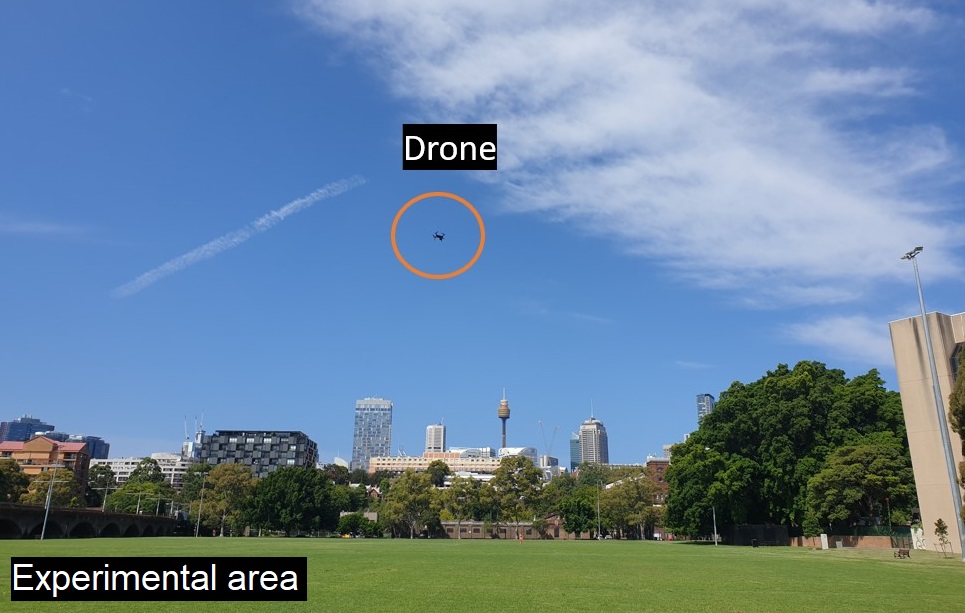}
    \caption{Experimental environment}
    \label{fig:FlyEnviroment}
\end{subfigure}
\begin{subfigure}[b]{.48\textwidth}
    \centering
    \includegraphics[width=\textwidth]{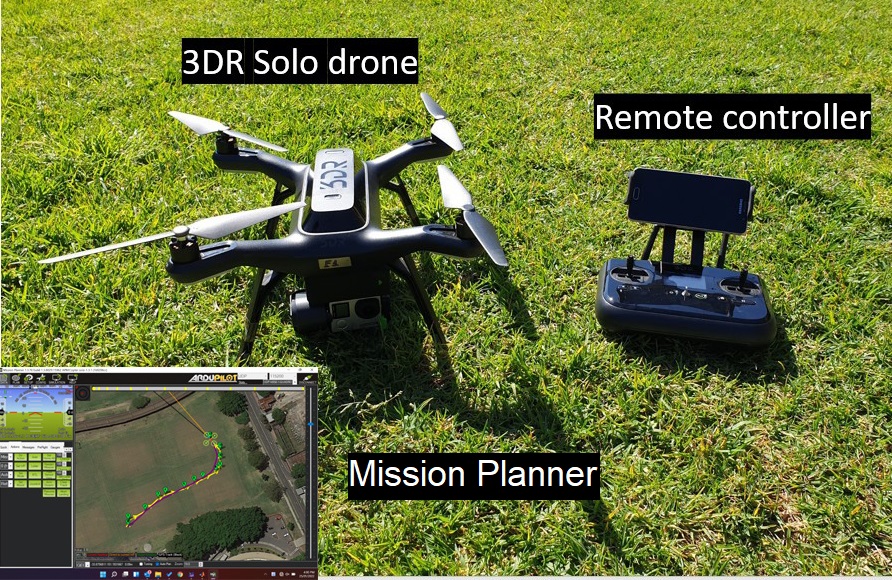}
    \caption{3DR Solo drone and Mission Planner}
    \label{fig:Drone}
\end{subfigure}
\caption{The UAV used and experimental environment}
\label{fig:Drone_n_Enviroment}
\end{figure}

Comparisons with other popular metaheuristic algorithms, including the differential evolution (DE), genetic algorithm (GA), and artificial bee colony (ABC), have been conducted to further evaluate the performance of the NMOPSO. Those algorithms are implemented based on \cite{DE}, \cite{GAcomparePSO}, and \cite{ABC}, respectively. The generated paths are shown in Figure \ref{fig:MetaheuristicALMap1} and \ref{fig:MetaheuristicALMap2}. It can be seen that all algorithms are able to generate feasible paths. The NMOPSO, however, introduces the shortest paths with small turning angles. Table \ref{tbl:comparision_table1} shows the values of the objective functions corresponding to the paths generated by those algorithms. It can be seen that the ABC gives average results across all objectives. The DE produces good results for $F_2$ and $F_3$, but the path length is not optimized. The GA produces poor results in complex scenarios due to its node reduction mechanism. The NMOPSO introduces the best overall performance reflected via its lowest cost for $F_1$ and $F_4$ in most scenarios. The cost associated with the flight altitude is not as good as the DE due to kinematic constraints. However, it ensures the paths generated are feasible for the UAV to follow. In addition, the generation of multiple non-dominated solutions gives the NMOPSO capabilities to fulfill different needs of applications.

\subsection{Comparison with other multi-objective optimization algorithms}

Comparisons with other multi-objective optimization algorithms, including the original multi-objective particle swarm optimization (MOPSO) algorithm \mbox{\cite{1004388}}, non-dominated sorting genetic algorithm (NSGA-II) \mbox{\cite{996017}}, and Pareto envelope-based selection algorithm II (PESA-II) \mbox{\cite{PESA2}}, have been conducted to evaluate the non-dominated solutions generated by the NMOPSO. Metrics used for comparisons include the maximum value (max), minimum value (min), mean value (mean), and standard deviation (std) of each objective on the obtained Pareto front. A new metric called solution distribution, $s_d$, is also used to evaluate the diversity of non-dominated solutions. It is defined as:
\begin{equation}
\label{eqn:popDensity}
    s_d =\frac{n_p}{n_o},
\end{equation}
where $n_p$ is the number of non-dominated solutions found and $n_o$ is the number of cells occupied by those non-dominated solutions. A small value of $s_d$ thus indicates a good distribution of solutions, which is expected for the multi-objective problem.

\begin{sidewaystable}
\caption{Comparison result for multi-objective algorithms}
\label{tbl:MoComparisionTable}
\centering
% \resizebox{\textwidth}{!}{
\begin{tabular}{C{1cm}C{1.2cm}C{0.8cm}C{0.8cm}C{0.8cm}C{0.8cm}C{0.8cm}C{0.8cm}C{0.8cm}C{0.8cm}C{0.8cm}C{0.8cm}C{0.8cm}C{0.8cm}C{0.8cm}C{0.8cm}C{0.8cm}C{0.8cm}C{0.8cm}}
\hline \hline
Scens. &  Alg.   & \multicolumn{4}{c}{F1} & \multicolumn{4}{c}{F2} & \multicolumn{4}{c}{F3} & \multicolumn{4}{c}{F4} & {$s_d$} \\
          &         & Max    & Min    & Mean            & Std    & Max    & Min    & Mean            & Std    & Max    & Min    & Mean            & Std    & Max    & Min    & Mean            & Std    &                     \\ \hline \hline
\multirow{4}{*}{1}         & NMOPSO  & 0.0615 & 0.0107 & \textbf{0.0178} & 0.0099 & 0.0434 & 0.0000 & 0.0141          & 0.0119 & 0.6742 & 0.0642 & 0.2691          & 0.1621 & 0.0663 & 0.0243 & \textbf{0.0376} & 0.0092 & \textbf{1.1354}       \\
          & NSGA-II & 0.3104 & 0.0325 & 0.1225          & 0.0802 & 0.0500 & 0.0000 & 0.0068          & 0.0091 & 0.7786 & 0.0172 & 0.2654          & 0.2391 & 0.1683 & 0.0690 & 0.1028          & 0.0247 & 2.2644                \\
          & PESA-II & 0.6089 & 0.5317 & 0.5630          & 0.0232 & 0.0098 & 0.0000 & \textbf{0.0020} & 0.0028 & 0.2639 & 0.1755 & \textbf{0.2120} & 0.0257 & 0.5047 & 0.3566 & 0.4189          & 0.0447 & 2.8368                \\
          & MOPSO   & 0.3734 & 0.2035 & 0.2826          & 0.0408 & 0.0306 & 0.0000 & 0.0087          & 0.0083 & 0.1632 & 0.0071 & 0.0550          & 0.0485 & 0.2765 & 0.1308 & 0.1771          & 0.0350 & 1.1831                \\ \hline
\multirow{4}{*}{2}         & NMOPSO  & 0.0472 & 0.0249 & \textbf{0.0314} & 0.0050 & 0.0293 & 0.0000 & 0.0092          & 0.0085 & 0.6585 & 0.0869 & 0.2741          & 0.1464 & 0.0551 & 0.0289 & \textbf{0.0372} & 0.0060 & \textbf{1.0861}       \\
          & NSGA-II & 0.2918 & 0.0569 & 0.1464          & 0.0671 & 0.0421 & 0.0000 & 0.0058          & 0.0081 & 0.7905 & 0.0184 & 0.2325          & 0.2410 & 0.1888 & 0.0721 & 0.1158          & 0.0331 & 2.2087                \\
          & PESA-II & 0.6254 & 0.5711 & 0.5856          & 0.0130 & 0.0026 & 0.0000 & \textbf{0.0006} & 0.0008 & 0.2505 & 0.1959 & 0.2159 & 0.0134 & 0.4146 & 0.3286 & 0.3548          & 0.0234 & 3.3209                \\
          & MOPSO   & 0.2085 & 0.1880 & 0.1955          & 0.0049 & 0.0314 & 0.0000 & 0.0089          & 0.0083 & 0.0475 & 0.0139 & \textbf{0.0274}          & 0.0089 & 0.1416 & 0.1075 & 0.1185          & 0.0070 & 1.0865                \\ \hline
\multirow{4}{*}{3}         & NMOPSO  & 0.0759 & 0.0119 & \textbf{0.0234} & 0.0143 & 0.0388 & 0.0000 & 0.0109          & 0.0106 & 0.6107 & 0.0748 & 0.2947          & 0.1596 & 0.0860 & 0.0306 & \textbf{0.0449} & 0.0118 & \textbf{1.1827}       \\
          & NSGA-II & 0.2930 & 0.0505 & 0.1298          & 0.0661 & 0.0435 & 0.0000 & 0.0062          & 0.0083 & 0.7436 & 0.0342 & 0.2277 & 0.2110 & 0.2011 & 0.0697 & 0.1168          & 0.0361 & 2.2316                \\
          & PESA-II & 0.4817 & 0.4442 & 0.4648          & 0.0114 & 0.0059 & 0.0000 & \textbf{0.0022} & 0.0019 & 0.2863 & 0.1905 & 0.2514          & 0.0290 & 0.4012 & 0.3503 & 0.3763          & 0.0166 & 3.0891                \\
          & MOPSO   & 0.2276 & 0.1752 & 0.1911          & 0.0146 & 0.0188 & 0.0000 & 0.0049          & 0.0046 & 0.0629 & 0.0119 & \textbf{0.0316}          & 0.0129 & 0.1420 & 0.0709 & 0.0933          & 0.0187 & 1.2853                \\ \hline
\multirow{4}{*}{4}         & NMOPSO  & 0.3204 & 0.0797 & \textbf{0.1619} & 0.0593 & 0.0303 & 0.0000 & 0.0067          & 0.0079 & 0.7717 & 0.0773 & 0.3245          & 0.1875 & 0.1626 & 0.0695 & \textbf{0.0975} & 0.0210 & \textbf{1.1567}       \\
          & NSGA-II & 0.3981 & 0.1075 & 0.2419          & 0.0715 & 0.0342 & 0.0000 & \textbf{0.0040} & 0.0062 & 0.8137 & 0.0436 & 0.2889          & 0.2288 & 0.2463 & 0.0843 & 0.1452          & 0.0438 & 2.2546                \\
          & PESA-II & 0.5497 & 0.4952 & 0.5223          & 0.0179 & 0.0047 & 0.0000 & 0.2311          & 0.0015 & 0.2696 & 0.1944 & 0.2311          & 0.0228 & 0.4366 & 0.3500 & 0.3854          & 0.0274 & 3.0496                \\
          & MOPSO   & 0.2745 & 0.1848 & 0.2367          & 0.0245 & 0.0252 & 0.0000 & 0.0072          & 0.0068 & 0.2239 & 0.0240 & \textbf{0.0776} & 0.0496 & 0.1828 & 0.0890 & 0.1223          & 0.0221 & 1.2374      \\ \hline\hline          
\end{tabular}
% }
\end{sidewaystable}

The comparison result is shown in Table \ref{tbl:MoComparisionTable}. It can be seen that the proposed NMOPSO outperforms other algorithms in objectives $F_1$ and $F_4$. Particularly in objective $F_1$, NMOPSO demonstrates the best optimization capability across all three metrics: max, min, and mean. For objectives $F_2$ and $F_3$, PESA-II yields good mean values. However, when considering the max, min, and std metrics in scenarios 2, 3, and 4, along with the result for other objectives, it is evident that PESA-II produces less diverse solutions. This is also reflected in its highest $s_d$ value. NSGA-II provides reasonable results but is not dominant in any particular objective. Similarly, MOPSO only achieves average results due to the local minimum issue. NMOPSO resolves this issue through the proposed adaptive mutation mechanism. This mechanism also enhances the solution distribution reflected via the best $s_d$ values of NMOPSO in most scenarios.

\subsection{Experimental validation}
\begin{figure}
\begin{subfigure}{.46\textwidth}
    \centering
    \includegraphics[width=\textwidth]{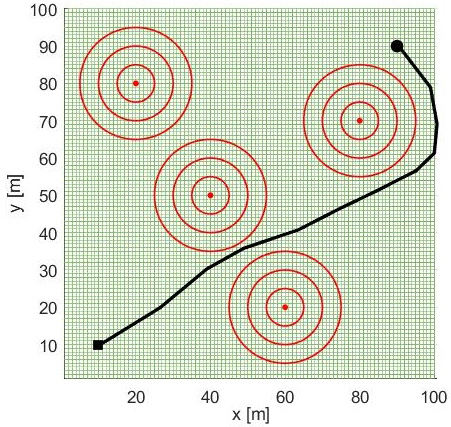}
    \caption{Top view of the planned path generated by NMOPSO in MATLAB}
    \label{fig:ExperimentResultxy}
\end{subfigure}
\begin{subfigure}{.49\textwidth}
    \centering
    \includegraphics[width=0.95\textwidth]{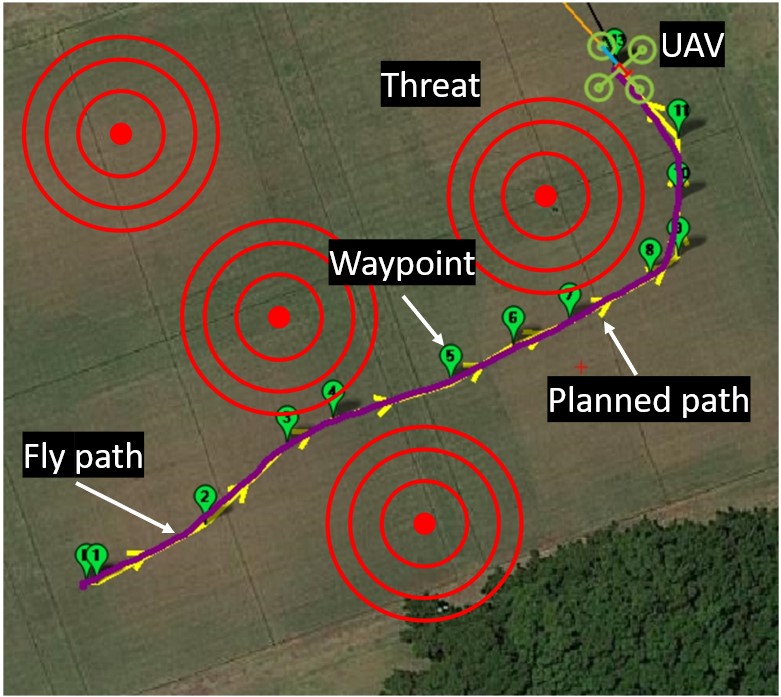}
    \caption{The planned path (yellow) and the actual flight  paths (purple) in experiments}
    \label{fig:ExperimentResultz}
\end{subfigure}
\caption{Experimental results}
\label{fig:ExperimentResult}
\end{figure}

To verify the validity of the proposed algorithm in generating paths for practical flights, experiments have been carried out with a real UAV named 3DR Solo. This UAV can be programmed to fly automatically via a ground control station software called Mission Planner, as depicted in Figure \mbox{\ref{fig:Drone}}. The experimental area is located on a flat terrain at the latitude and longitude of $(-33.876399, 151.192293)$. It has the size of $100 \times 100$ m$^2$ and is augmented with four obstacles, as shown in Figure \mbox{\ref{fig:FlyEnviroment}}. This information, along with the UAV's starting and goal locations, is input into the NMOPSO algorithm implemented in MATLAB to generate a planned path comprising a list of waypoints, as shown in Figure \mbox{\ref{fig:ExperimentResultxy}}. Those waypoints are then converted to geographic coordinates and uploaded to the 3DR Solo drone via Mission Planner to fly.

Figure \ref{fig:ExperimentResultz} shows the flight result in which the yellow line represents the planned path and the purple line represents the actual flight path. Their overlap implies that the path generated by NMOPSO is feasible for the drone to follow. Since the flight path does not intersect the threat circles, the path is safe for drone operation. The results thus demonstrate the validity of NMOPSO for practical flights.

\subsection{Discussion}
Using a multi-objective approach, the NMOPSO generates a set of non-dominated solutions that can meet various application requirements. For example, the third solution for Scenario 2 in Table {\ref{tbl:comparision_table1}} minimizes the energy consumption as it represents a short and smooth path, while the first and second solutions prioritize safety. The NMOPSO also allows new objectives to be added as additional dimensions of its hypercube. The algorithm is, therefore, scalable and suitable for complex tasks. Besides, the inclusion of kinematic constraints to the problem helps narrow down the solution space. The NMOPSO takes advantage of this by utilizing navigation variables to speed up the process of finding non-dominated solutions. The multi-objective approach, however, faces the challenge of finding a large number of non-dominated solutions, which requires a balance between exploration and exploitation. Some of our enhancements to the NMOPSO, such as utilizing a repository to store non-dominated solutions and implementing a region-based mutation mechanism, can address this issue, but at the cost of adding additional computation requirements. 

\section{Conclusion} \label{sec:con}
In this work, a new path-planning algorithm, NMOPSO, has been introduced to generate Pareto optimal paths for UAVs considering their kinematic constraints. Several mechanisms such as navigation variables, fitness evaluation, and adaptive mutation have been integrated into the algorithm to better explore the solution space for non-dominated solutions. Comparison results show that the NMOPSO outperforms other PSO variants and state-of-the-art metaheuristic optimization algorithms in most criteria including path length, safety and smoothness. In addition, experiments with paths generated for real flights have been conducted. The overlap between the planned and actual flight paths confirms the validity of our approach for practical UAV operations.

% \subsection*{Author contributions}
% Thi Thuy Ngan Duong: Conceptualization, Methodology, Implementation, Writing - original draft. \\
% Duy-Nam Bui: Methodology, Writing - original draft.\\
% Manh Duong Phung: Conceptualization, Investigation, Writing - review and editing, Supervision.

\subsection*{Conflict of interest} On behalf of all authors, the corresponding author states that there is no conflict of interest.

\subsection*{Data availability statement}
The data generated in this study is available from the authors on reasonable request.

\subsection*{Financial interests}
The authors have no relevant financial or non-financial interests to disclose.

\bibliography{mybibfile}

\end{document}